\documentclass{article}

\usepackage{amsmath,amsfonts,bm}

\def\eqref#1{equation~\ref{#1}}

\def\1{\bm{1}}

\def\va{{\bm{a}}}
\def\vb{{\bm{b}}}
\def\vc{{\bm{c}}}
\def\vd{{\bm{d}}}
\def\ve{{\bm{e}}}

\def\vo{{\bm{o}}}

\def\vx{{\bm{x}}}
\def\vy{{\bm{y}}}

\DeclareMathAlphabet{\mathsfit}{\encodingdefault}{\sfdefault}{m}{sl}
\SetMathAlphabet{\mathsfit}{bold}{\encodingdefault}{\sfdefault}{bx}{n}

\def\sA{{\mathbb{A}}}

\def\sD{{\mathbb{D}}}

\def\sL{{\mathbb{L}}}

\def\sP{{\mathbb{P}}}

\def\sU{{\mathbb{U}}}

\def\sX{{\mathbb{X}}}
\def\sY{{\mathbb{Y}}}
\def\sZ{{\mathbb{Z}}}

\newcommand{\R}{\mathbb{R}}

\DeclareMathOperator*{\argmax}{arg\,max}

\DeclareMathOperator{\sign}{sign}

\usepackage{listings}

\usepackage{booktabs}
\usepackage{wrapfig}
\usepackage{multirow}
\usepackage{hhline}
\usepackage{colortbl}
\usepackage{microtype}
\usepackage{graphicx}
\usepackage{subcaption}
\usepackage{arydshln}
\usepackage{anyfontsize}

\usepackage{booktabs} %

\usepackage{hyperref}
\usepackage{booktabs} %
\usepackage{amsmath,amssymb,amsthm}
\usepackage{boxedminipage}
\usepackage{stmaryrd}
\usepackage{marvosym}
\usepackage{lmodern}
\usepackage{tikz}

\usepackage{floatrow}

\usepackage{cuted}

\usepackage{pgfplots}
\usepackage{pgfplotstable}
\usepackage{xcolor}
\usepackage{tabu}

\usepackage[none]{hyphenat}

\newcommand{\fig}[1]{Fig.~\ref{#1}}

\usetikzlibrary{quotes,angles,positioning,calc, shapes, arrows, 3d, decorations.pathreplacing,angles,quotes}

\definecolor{darkred}{rgb}{0.5,0,0}
\definecolor{darkgreen}{rgb}{0,0.5,0}
\definecolor{darkblue}{rgb}{0,0,0.5}
\definecolor{lightgrey}{rgb}{0.7,0.7,0.7}

\definecolor{lightergrey}{rgb}{0.93,0.93,0.93}

\usepackage{placeins}
\usepackage{fancyvrb}

\usepackage[accepted]{icml2020}

\theoremstyle{definition}

\newcommand{\domain}{\textsc{ApproxLine}}
\newcommand{\eline}{\textsc{Exact}}

\newcommand{\aedcyclegan}{CycleAE}
\newcommand{\vae}{VAE}

\begin{document}

\icmltitlerunning{Robustness Certification of Generative Models}
\twocolumn[
\icmltitle{Robustness Certification of Generative Models}

\begin{icmlauthorlist}
\icmlauthor{Matthew Mirman}{to}
\icmlauthor{Timon Gehr}{to}
\icmlauthor{Martin Vechev}{to}
\end{icmlauthorlist}

\icmlaffiliation{to}{Department of Computer Science, ETH Zurich, Zurich, Switzerland}

\icmlcorrespondingauthor{Matthew Mirman}{matthew.mirman@inf.ethz.ch}

\icmlkeywords{Machine Learning, ICML}
\vskip 0.3in
]

\printAffiliationsAndNotice{}  %

\setlength{\abovedisplayskip}{3pt}
\setlength{\belowdisplayskip}{3pt}

\begin{abstract}
Generative neural networks can be used to specify continuous transformations between images via latent-space interpolation. However, certifying that all images captured by the resulting path in the image manifold satisfy a given property can be very challenging. This is because this set is highly non-convex, thwarting existing scalable robustness analysis methods, which are often based on convex relaxations. 
We present \domain{}, a scalable certification method that successfully verifies non-trivial specifications involving generative models and classifiers. \domain{} can provide both sound deterministic and probabilistic guarantees, by capturing either infinite non-convex sets of neural network activation vectors or distributions over such sets. We show that \domain{} is practically useful and can verify interesting interpolations in the network’s latent space.
\end{abstract}

\section{Introduction}

While neural networks are used across a wide range of applications, from facial recognition to autonomous driving, certification of their behavior has remained predominantly focused on uniform classification of norm-bounded balls aiming to capture invisible perturbations \citep{ai2,katz2017reluplex,wong2018scaling, ibp, eran, raghunathan2018certified, tjeng2017evaluating, dvijotham2018training, salman2019convex, dvijotham2018dual, wang2018efficient}.

However, a system's safety can depend on its behavior on visible transformations. Consequently, there is increased interest in investigating techniques able to certify more complex specifications (e.g., geometric perturbations) \citep{balunovic2019geometric, liu2019certifying,  dvijotham2018verification, singh2019abstract}.

Of particular interest is the work of \citet{exactline}, which shows that if inputs of a network are restricted to a line segment, exact verification can sometimes be efficiently performed. This method has been used to certify properties which are not norm-based \citep{acas} and improve Integrated Gradients \citep{sundararajan2017axiomatic}.

\begin{figure}[t]
\centering
\begin{tikzpicture}[
mnode/.style={circle,draw=black,fill=black,inner sep=0pt,minimum size=2pt}
]

\draw (-3,-0.5) node[mnode]{}
   -- (-1.5,0.5) node[mnode]{}
   -- (0,0.7) node[mnode]{}
   -- (1.5,0.5) node[mnode]{}
   -- (3,-0.5) node[mnode]{};

\draw[red, dashed] (-3,-0.5) node[mnode]{}
   -- (0,-0.5) node[mnode]{}
   -- (3,-0.5) node[mnode]{};

\node[inner sep=0pt] at (-3,-0.5)
    {\includegraphics[width=0.975cm]{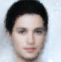}};

\node[inner sep=0pt] at (-1.5,0.5)
    {\includegraphics[width=0.975cm]{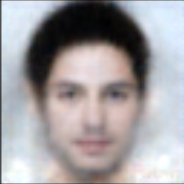}};

\node[inner sep=0pt] at (0,0.7)
    {\includegraphics[width=0.975cm]{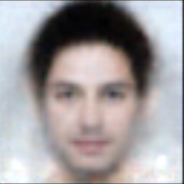}};

\node[inner sep=0pt] at (1.5,0.5)
    {\includegraphics[width=0.975cm]{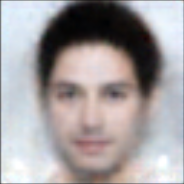}};

\node[inner sep=0pt] at (3,-0.5)
    {\includegraphics[width=0.975cm]{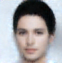}};

\node[inner sep=0pt] at (0,-0.5)
    {\includegraphics[width=0.975cm]{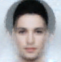}};

\end{tikzpicture}
\caption{To prove that a classifier is robust to orientation changes, we consider a non-convex set of images created by interpolation in the latent space of a generative model. The bottom-middle image is the nonsensical pixel-wise average of the endpoints that every convex relaxation based analysis would have to consider.}
\label{introfig}
\vspace{-0.16in}
\end{figure}

\vspace{-0.05in}
\paragraph{This Work} In this work we build on related ideas and introduce a powerful verifier based on non-convex overapproximation of sets of neuron activation vectors, trading off some exactness for scalability. Further, our verifier is able to certify \emph{probabilistic} properties, which can be used to quantify the fraction of inputs that satisfy the desired property even if it does not hold for all considered inputs. In contrast to sampling-based approaches, our analyzer computes guaranteed bounds on those quantities.

\begin{figure*}[t]
\centering

\begin{tikzpicture}

\fill[yellow!10] (-9.95,1.2) rectangle (-8.5,-3.8);
\node at (-9.225,1.5) {Encode};

\fill[black!5] (-8.5,1.2) rectangle (-6.1,-3.8);
\node at (-7.3,1.5) {Decode};

\fill[green!7] (-6.1,1.2) rectangle (6.7,-3.8);
\node at (0.3,1.5) {Classify with $n_A$};

  \draw (-8,0.5)  circle (0.35cm) node[align=center,anchor=center] (X01) {$E_1$};
  \draw (-8,-0.5) circle (0.35cm) node[align=center,anchor=center] (X02) {$E_2$};

  \draw[decoration={brace,raise=0.4cm, amplitude=0.17cm },decorate] ($(X01) + (0,0.3)$) -- node [anchor=west, midway, xshift=0.35cm](X0) {} ($(X02) - (0,0.3)$);

  \draw (-5.5,0.5)  circle (0.35cm) node[align=center,anchor=center] (X11) {$x_{1,1}$};
  \draw (-5.5,-0.5) circle (0.35cm) node[align=center,anchor=center] (X12) {$x_{1,2}$};

  \draw[decoration={brace,mirror,raise=0.4cm, amplitude=0.17cm},decorate]($(X11) + (0,0.3)$) -- node [anchor=east, midway, xshift=-0.35cm] (X1) {} ($(X12) - (0,0.3)$);

  \draw [->, densely dotted] (X0) -- node[midway, above] {$n_D$} (X1);

  \draw (-0.36,0.5)  circle (0.35cm) node[align=center,anchor=center] (X21) {$x_{2,1}$};
  \draw (-0.36,-0.5) circle (0.35cm) node[align=center,anchor=center] (X22) {$x_{2,2}$};

  \draw [->] (X11) -- node[above] {ReLU} (X21);
  \draw [->] (X12) -- node[above] {ReLU} (X22);

  \draw (4.5,0.5)  circle (0.35cm) node[align=center,anchor=center] (X31) {$x_{3,1}$};
  \draw (4.5,-0.5) circle (0.35cm) node[align=center,anchor=center] (X32) {$x_{3,2}$};

  \draw [->] (X21) -- node[midway, above = -0.07] {\small $0.5$} (X31);
  \draw [->] (X22) -- node[pos=0.2, above = -0.07, sloped] {\small $0.5$} (X31);

  \draw [->] (X21) -- node[pos=0.35, above = -0.07, sloped] {\small $1$} (X32);
  \draw [->] (X22) -- node[midway, above = -0.07] {\small $-0.25$} (X32);

\begin{axis}[
width=1.32cm,
height=2.2cm,
xshift=-8.4cm,
yshift=-3.75cm,
scale only axis,
ticklabel style={font=\tiny},
xtick={-1,1,2},
ytick={1,2,3,4},
xmin=-1.2,
xmax=3,
ymin=-1.2,
ymax=5,
axis x line=center,axis y line=center,
axis x line shift=0,
axis y line shift=0,
xlabel={$E_{1}$},
ylabel={$E_{2}$},
xlabel style={at=(current axis.right of origin), anchor=west},
ylabel style={at=(current axis.above origin), anchor=south},
]

\addplot [color=blue!60,,mark=*,mark size=1.5pt, forget plot] coordinates { (1.5,4) (2.2,1.5) }
    node[pos=0,above=-0.1mm] (E1) {\color{black}$\ve_1$}
    node[pos=0.5,above=-0.6mm, sloped] {\tiny\color{red}$\lambda = 1$}
    node[pos=1,below=0.1mm] (E2) {\color{black}$\ve_2$};

\node (graph_0_right) at (axis cs:3,2.5){};

\end{axis}
\node at (-7.75, -4.1) {(a)};

\node[inner sep=0pt] (P1) at ($(E1) - (2,0.2)$)
    {\includegraphics[width=0.7cm]{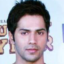}};

\node[inner sep=0pt] (P2) at ($(P1) - (0,1.4)$)
    {\includegraphics[width=0.7cm]{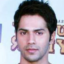}};

\draw[->, densely dotted] (P1) to node[pos=0.3, above] {$n_E$} (E1);
\draw[->, densely dotted] (P2) -- node[pos=0.2, above] {$n_E$} (E2);

\begin{axis}[
width=1.845cm,
height=2.2cm,
xshift=-6cm,
yshift=-3.75cm,
scale only axis,
ticklabel style={font=\tiny},
xtick={-1,1,2,3},
ytick={1,2,4},
xmin=-1.2,
xmax=4.5,
ymin=-1.2,
ymax=5,
axis x line=center,axis y line=center,
axis x line shift=0,
axis y line shift=0,
xlabel={$x_{1,1}$},
ylabel={$x_{1,2}$},
xlabel style={at=(current axis.right of origin), anchor=west},
ylabel style={at=(current axis.above origin), anchor=south},
]

\addplot [color=blue!60,,mark=*,mark size=1.5pt, forget plot] coordinates { (1,2) (-1,3) (-1,3.5) (1,4.5) (3.5,2) }
    node[pos=0, inner sep=0pt] (D1) {}
    node[pos=0.06,above=-0.6mm, sloped] {\tiny\color{red}$.2$}
    node[pos=0.3,right=-0.6mm] {\tiny\color{red}$.2$}
    node[pos=0.5,below=-0.6mm, sloped] {\tiny\color{red}$.2$}
    node[pos=0.75,above=-0.6mm, sloped] {\tiny\color{red}$.4$}
    node[pos=1] (D2) {};

\node (graph_a_left) at (axis cs:-1.5,2.5){};
\node (graph_a_right) at (axis cs:4,2.5){};

\end{axis}
\node at (-5.1, -4.1) {(b)};

\draw[->, densely dotted] (graph_0_right) -- node[midway, above] {$n^\#_D$} (graph_a_left);

\node[inner sep=0pt] (D1I) at ($(D1) + (0.0,-0.32)$) {\includegraphics[width=0.4cm]{images/flip_n_left.png}};
\node[inner sep=0pt] (D2I) at ($(D2) + (-0.15,-0.32)$) {\includegraphics[width=0.4cm]{images/flip_n_right.png}};

\begin{axis}[
width=1.74cm,
height=2.2cm,
xshift=-2.75cm,
yshift=-3.75cm,
scale only axis,
ticklabel style={font=\tiny},
xtick={1,2,3},
ytick={1,2,4},
xmin=-0.9,
xmax=4,
ymin=-1.2,
ymax=5,
axis x line=center,axis y line=center,
axis x line shift=0,
axis y line shift=0,
xlabel={$x_{2,1}$},
ylabel={$x_{2,2}$},
xlabel style={at=(current axis.right of origin), anchor=west},
ylabel style={at=(current axis.above origin), anchor=south},
]
\addplot [color=blue!60,,mark=*,mark size=1.5pt, forget plot] coordinates { (1,2) (0,2.5) (0,3) (0,3.5) (0,4) (1,4.5) (3.5,2) }
    node[pos=0.05,below=-0.6mm, sloped] {\tiny\color{red}$.1$}
    node[pos=0.2,right=-0.6mm] {\tiny\color{red}$.1$}
    node[pos=0.25,left=-0.6mm] {\tiny\color{red}$.2$}
    node[pos=0.33,right=-0.6mm] {\tiny\color{red}$.1$}
    node[pos=0.445,above=-0.6mm, sloped] {\tiny\color{red}$.1$}
    node[pos=0.7,above=-0.6mm, sloped] {\tiny\color{red}$.4$};

\node (graph_b_left) at (axis cs:-1.5,2.5){};
\node (graph_b_right) at (axis cs:4,2.5){};
\end{axis}
\node at (-1.9, -4.1) {(c)};

\draw[->] (graph_a_right)-- node[above]{ReLU$^\#$} (graph_b_left);

\begin{axis}[
width=1.74cm,
height=2.2cm,
xshift=0.5cm,
yshift=-3.75cm,
scale only axis,
ticklabel style={font=\tiny},
xtick={1,2,3},
ytick={1,2,3,4},
xmin=-1,
xmax=4,
ymin=-1.2,
ymax=5,
axis x line=center,axis y line=center,
axis x line shift=0,
axis y line shift=0,
xlabel={$x_{2,1}$},
ylabel={$x_{2,2}$},
xlabel style={at=(current axis.right of origin), anchor=west},
ylabel style={at=(current axis.above origin), anchor=south},
]
\addplot [color=blue!60,,mark=*,mark size=1.5pt, forget plot] coordinates { (1,2) (0,2.5) (0,3) (0,3.5) (0,4) (1,4.5) (3.5,2) }
    node[pos=0.7,above=-0.6mm, sloped] {\tiny\color{red}$.4$};

\addplot [draw=none, fill=orange!90, fill opacity=0.7, forget plot] coordinates { (0, 2) (0, 4.5) (1, 4.5) (1, 2) (0, 2) }
\closedcycle;

\node at (axis cs:1.35,3){\tiny\color{red} $.6$};

\node (graph_c_left) at (axis cs:-1.5,2.5){};
\node (graph_c_right) at (axis cs:4,2.5){};
\end{axis}
\node at (1.4, -4.1) {(d)};

\draw[->] (graph_b_right)-- node[above]{Relax} (graph_c_left);

\draw[decoration={brace,raise=37pt, amplitude=0.25cm},decorate](graph_b_left) -- (graph_c_right);

\begin{axis}[
width=1.756cm,
height=2.2cm,
xshift=4.25cm,
yshift=-3.75cm,
scale only axis,
ticklabel style={font=\tiny},
xtick={3},
ytick={1,2,3,4},
xmin=-1.2,
xmax=3.75,
ymin=-1.2,
ymax=5,
axis x line=center,axis y line=center,
axis x line shift=0,
axis y line shift=0,
xlabel={$x_{3,1}$},
ylabel={$x_{3,2}$},
xlabel style={at=(current axis.right of origin), anchor=west},
ylabel style={at=(current axis.above origin), anchor=south},
]
\addplot [color=blue!60,,mark=*,mark size=1.5pt, forget plot] coordinates { (1.5, 0.5) (1.25, -0.625) (1.5, -0.75) (1.75, -0.875) (2.0, -1.0) (2.75, -0.125) (2.75, 3)  }
    node[pos=0.8,right=-0.6mm] {\tiny\color{red}$.4$};

\addplot [draw=none, fill=orange!90, fill opacity=0.7, forget plot] coordinates { (1, -1.125) (1, 0.5) (2.75, 0.5) (2.75, -1.125) (1, -1.125) }
\closedcycle;

\node at (axis cs:2.15,0.85){\tiny\color{red} $.6$};

\node (graph_d_left) at (axis cs:-1.5,2.5){};
\node (graph_d_right) at (axis cs:5,2.5){};

\end{axis}
\node at (5.1, -4.1) {(e)};

\draw[->] (graph_c_right)-- node[above]{MatMul$^\#$} (graph_d_left);

\end{tikzpicture}
\caption{
Using \domain{} to find probability bounds for latent space interpolation of flipped images, as summarized in the pseudocode in Appendix~\ref{alg}.
Blue polygonal chains represent activation distributions at each layer exactly.
The orange boxes represent the relaxation that \domain{} would create,
obviating the need to keep track of the segments it entirely covers.
The probabilities associated with each segment or box are shown in red.
While in practice, images are more than two dimensional, and the classifier we use is much larger, the presented inference presented in the lightly shaded green region is faithful to the weights of the network shown in the top row.
}

\label{abstract_example}
\end{figure*}

\paragraph{Main contributions} Our key contributions are:
\begin{itemize}
\item A verification system, \domain{}\footnote{ \href{https://www.dropbox.com/s/np89rh2q8hzr1pj/approxline_submit.tar.gz?dl=0}{Link to code and model snapshots in dropbox.}}, which introduces sound approximations to non-convex certification, scaling to much larger networks (nearly 100k neurons) than exact methods and even sampling.
\item The first demonstration of verification of visible specifications based on latent space interpolations of a generator, such as that a classifier for ``is bald'' is robust to different orientations of a head, as in \fig{introfig}.
\item A method to compute tight deterministic \emph{guaranteed bounds} on probabilities of outputs given distributions over inputs, which we believe is the first application of probabilistic abstract interpretation in the context of neural networks.
\end{itemize}

\paragraph{Related work}

\citet{dvijotham2018verification} verify probabilistic properties universally over sets of inputs by bounding the probability that a dual approach verifies the property. In contrast, our system verifies properties that are either universally quantified or probabilistic. However, the networks we verify are multiple orders of magnitude larger. While they only provide upper bounds on the probability that a specification has been violated, we provide extremely tight bounds on such probabilities from both sides. 
PROVEN \citep{weng2018proven} proposes a technique to infer confidence intervals on the probability of misclassification from prexisting convex relaxation methods that find linear constraints on outputs.
We show how in the case of interpolations for generative models, convex relaxation methods are unable to prove meaningful bounds. This implies that the linear lower bound function used by PROVEN would be bounded above by 0, and thus because $F_{g_t}^L(0.5) \geq F_{0}^L(0.5)$ and $F_{0}^L(0.5) = 1$, the lower bound, $\gamma_L$, that their system should derive would be $\gamma_L = 0$.  This is because even the most precise convex relaxation over the generated set of images might include many nonsensical images. For example, the convex hull includes the pixel-wise average of the generated endpoint images, as can be seen in \fig{introfig}.

Another line of work is smoothing, which provides a defense with statistical robustness guarantees \citep{cohen2019certified, lecuyer2018certified, liu2018towards, li2018second, cao2017mitigating}. In contrast, \domain{} provides deterministic guarantees, and is not a defense. The analysis we use to determine bounds is an instance of probabilistic abstract interpretation \citep{probAI}.

\section{Overview of \domain{}}
We now use the example in \fig{abstract_example} to demonstrate how \domain{} computes exact and probabilistic bounds for the robustness of a classifier based on a latent space image transformation.
The goal is to verify that a classification network $n_A$ does not change its prediction when presented with images of a head from different angles, produced by interpolating encodings in the latent space of an autoencoder.
We first use the encoder, $n_E$, to produce encodings $\ve_1$ and $\ve_2$ from the original image and that image flipped horizontally.
It is a common technique to use the decoder $n_D$ to get a picture of a head at an intermediate angle, on an interpolated point $\ve$, taken from the segment $\overline{\ve_1\ve_2} =\{\ve_1+\alpha\cdot(\ve_2-\ve_1)~~|~~\alpha \in [0,1]\}$. Decodings for $\ve_1$ and $\ve_2$ can be seen in \fig{abstract_example}(b). We want to check the property for {\em all} possible encodings on $\overline{\ve_1\ve_2}$ (not only points $\ve_1$ and $\ve_2$).

To accomplish this, we propagate lists of line segments and interval (box) constraints through the decoder and classifier, starting with the segment $\overline{\ve_1\ve_2}$.
At each layer, we adaptively relax this list by combining segments into interval constraints, in order to reduce the number of points that need to be managed in downstream layers.
This relaxation is key, as without it, the number of tracked points could grow exponentially with the number of layers.
While \cite{exactline} demonstrates that this is not a significant concern when propagating through classifiers, for generative models or decoders, the desired output region will be highly non-convex (with better models producing more segments). One may think of the number of segments produced by the model in such a case as the model's ``generative resolution''.

\textbf{Example of Inference with Overapproximation} Consider the simple (instructive) two dimensional input classifier network shown in \fig{abstract_example}, with inputs $x_{1,1}$ and $x_{1,2}$.
The possible inputs to this network we would like to consider are the points in the region described by the blue polygonal chain in \fig{abstract_example}(b), whose axes are $x_{1,1}$ and $x_{1,2}$.
The chain has coordinates $(1,2), (-1,3), (-1,3.5), (1,4.5), (3.5,2)$ with  $(1,2)$ representing $n_D(\ve_1)$ and $(3.5,2)$ representing $n_D(\ve_2)$.
The segments of the chain are annotated with weights $\lambda = 0.2, 0.2, 0.2, 0.4$.
These weights are such that the distribution produced by picking segment $j$ with probability $\lambda^{(j)}$ and then picking a point uniformly on that segment is the same as the distribution of $n_D(\ve)$ given $\ve \sim U(\overline{\ve_1\ve_2})$ where $U(S)$ is the uniform distribution on S.

After applying the ReLU layer to this chain (marked with ReLU$^\#$), one can observe in \fig{abstract_example}(c) that the first and third segment of this chain are split in half, resulting in 6 segments, which is 50\% more than there were originally. As the segments represented uniform distributions, the weights of the new segments is the proportional weight of that part on the pre-ReLU segment.
Here, each part of the new segment obtains half the pre-ReLU segment's weight.

Since a 50\% increase is significant, we now decide to consolidate, moving from exact to approximate (still sound) inference.
Here, we use a heuristic (labeled Relax), to choose segments to subsume that are small and close together. As they are clearly quite close together, we pick the first $5$ segments,
replacing them by the (orange) box that has smallest corner at $(0,2)$ and largest corner at $(1,4.5)$.
This box, introduced in \fig{abstract_example}(d), is assigned a weight equivalent to the sum $0.6$, of the weights of all removed segments.
Whereas each segment represents a uniform distribution, this new box represents a specific but unknown distribution with all its mass in the box.
As a box is represented by two points (maximum and minimum or center and radius), only four points must be maintained, a significant reduction.

The last step is to perform matrix multiplication. As this is a linear transformation, segments can be transformed by transforming their nodes, without adding new segments.
Box constraints can be transformed using interval arithmetic, also without adding anything further.
The weights of the regions are preserved, as the probability of selecting each region has not changed, only the regions themselves.  

\textbf{Computing probabilistic bounds}
Let ${\sA'}^{(j)}$ for $j=1\ldots k$ represent the regions (either the box and segment, or all 6 segments) shown in \fig{abstract_example}(e), each with weight $\lambda^{(j)}$.  We want to find bounds on the probability of class $t = 1$ being selected by classifier $n_A$. We can immediately bound this probability $\Pr_{e \in \overline{\ve_1\ve_2}} [ \argmax_i n_D(e)_i = t ] $ from below by
\[
\sum_{j} [\forall x_3 \in {\sA'}^{(j)} . x_{3,1} > x_{3,2}] \lambda^{(j)}
\]
where $[P]$ is the indicator for predicate $P$. To bound it from above, we would compute analogously
\[
\sum_{j}[\exists x_3 \in {\sA'}^{(j)} . x_{3,1} > x_{3,2}] \lambda^{(j)}.
\]
As an example, we will compute the lower bound for the example where we used relaxations. Here, it is clear that the entirety of the orange box lies within the region where $x_{3,1} > x_{3,2}$, so its indicator is evaluated to $1$ and we use its weight.  On the other hand, the segment contains a point where $x_{3,1} = 2.75$ and $x_{3,2} = 3$ which violates this condition, so its indicator is evaluated to $0$ and its weight is not used.  We can thus show a probabilistic lower bound of $0.6$. Of course, because the segment represents a uniform distribution, we could in fact provide an exact lower bound in this case by computing the fraction of the segment that satisfies the condition.  We will describe this in more detail later. We now observe that all the regions which would have been preserved using the exact procedure (in blue) would have contributed the same amount to the lower bound, as they all entirely satisfy the constraint. Exact inference would produce the same lower bound, but uses 50\% more points. 

\section{Certification of Deterministic Properties}
We review concepts from prior work \citep{ai2} and formally define \domain{} for deterministic properties.

Our goal is to automatically show that images $\vx$ from a given set $\sX$ of valid inputs are mapped to desirable outputs from a set $\sY$. We can write this property as $f[\sX]\subseteq\sY$. For example, we can choose $f$ as a decoder, $\sX$ as a line segment in its latent space and $\sY$ as the set of images for which a given classifier detects a desired attribute.

Such properties compose naturally: If we want to show that $h[\sX]\subseteq\sZ$ for $h(\vx)=g(f(\vx))$, it suffices to find a set $\sY$ for which we can show $f[\sX]\subseteq\sY$ and $g[\sY]\subseteq\sZ$. For example, $f$ could be a decoder and $g$ an attribute detector, where $\sZ$ describes the set of output activations that lead to an attribute being detected.

We assume that we can decompose the neural network $f$ as a sequence of $l$ layers: $f=L_{l-1}\circ\cdots\circ L_0$. To show a property $f[\sX]\subseteq\sY$, we find a family of sets $\sA_0,\ldots,\sA_l$ such that $\sX\subseteq \sA_0$, $L_i[\sA_i]\subseteq\sA_{i+1}$ for $0\le i<l$ and $\sA_l\subseteq\sY$.

To automate this analysis, we choose $\sA_0,\ldots,\sA_l$ such that they can be represented using a few real parameters. We pick $\sA_0$ based on $\sX$ and $\sA_{i+1}$ based on $\sA_i$ and $L_i$, such that they satisfy the properties $\sX\subseteq \sA_0$ and $L_i[\sA_i]\subseteq\sA_{i+1}$. At the end, we check if we indeed have $\sA_l\subseteq\sY$. If this holds, the verification succeeds and we know that the property holds. Otherwise, the procedure fails to prove the property.

\paragraph{Interval Arithmetic} If we pick $\sA_0$ as a bounding box of $\sX$, we can compute sets $\sA_i$ for $1\le i\le l$ by evaluating the layers $L_i$ using interval arithmetic. The analysis computes a range of possible values for each activation of the neural network, i.e., the sets $\sA_i$ are boxes. At the end, we check if the bounds of $\sA_l$ are strong enough to place it inside $\sY$. This analysis is fast but imprecise --- it often fails to prove properties that actually hold.

\paragraph{Unions} We can also represent a set $\sA_i$ as a union of sets $\bigcup_{j=1}^{k_i}\sA_i^{(j)}$. If we know how to propagate the individual sets through layer $L_i$, we can (among other possibilities) independently propagate each set and form the union of the results to obtain $\sA_{i+1}$. For example, we can cover the set $\sX$ with boxes and then propagate them through the network independently using interval arithmetic for each box. In the end, we have to show that all resulting boxes are within $\sY$.

\subsection{\domain{} for Deterministic Properties}
\domain{} for deterministic properties represents the sets $\sA_i$ as unions of sets $\sA_i^{(j)}$, where each set $\sA_i^{(j)}$ is either a box or a line segment.
Let $n_i$ denote the number of neurons in layer $i$. Formally, $\sA_i=\bigcup_{j=1}^{k_i}\sA_i^{(j)}$, where for each $j$ either $\sA_i^{(j)}=\prod_{l=1}^{n_i}[a_l,b_l]$ is a box with given lower bounds $\va$ and upper bounds $\vb$, or $\sA_i^{(j)}=\overline{\vx_1\vx_2}$ is a  segment connecting given points $\vx_1$ and $\vx_2$ in $\R^{n_i}$. To automate the analysis, we can represent $\sA_i$ as a list of bounds of boxes and a list of pairs of endpoints of segments.

Like \textsc{ExactLine} \cite{exactline}, we focus on the case where the set $\sX$ of input activations is a segment. \textsc{ExactLine} discusses how to split a given segment into multiple segments that cover it, such that a given neural network is an affine function on each of the new segments. Essentially, \textsc{ExactLine} determines the points where the segment crosses decision boundaries of piecewise-linear activation functions and splits it at those points. In order to compute a set $\sA_{i+1}$ such that $L_i[\sA_i]\subseteq\sA_{i+1}$, we first split all segments according to this strategy applied to only the current layer $L_i$. Then, we map the endpoints of the resulting segments to the next layer by applying $L_i$ to all of them. This is valid and captures exactly the image of the segments under $L_i$, because due to the splits, $L_i$, restricted to any one of the segments, is always an affine function. Additionally, we propagate the boxes through $L_i$ by applying standard interval arithmetic.

Note that if we propagate a segment $\overline{x_1x_2}$ using this strategy alone for all layers, this analysis produces the exact image of $\sX$, which is essentially equivalent to performing the analysis using \textsc{ExactLine} exclusively.

\paragraph{Relaxation} Therefore, our analysis may, before applying a layer $L_i$, apply relaxation operators that turn the set $\sA_i$ into $\sA'_i$, such that $\sA_i\subseteq \sA'_i$. We use two kinds of relaxation operators: \emph{bounding box} operators remove a single segment $\overline{\vc\vd}$. The removed segment is replaced by its bounding box $\prod_{l=1}^{n_i}[\min(c_l,d_l),\max(c_l,d_l)]$. \emph{Merge} operators replace multiple boxes by their common bounding box.

By carefully applying the relaxation operators, we can explore a rich tradeoff between pure \textsc{ExactLine} and pure interval arithmetic. Our analysis generalizes both: if we never apply relaxation operators, the analysis reduces to \textsc{ExactLine}, and will be exact but potentially slow. If we relax the initial segment into its bounding box, the analysis reduces to interval arithmetic and will be imprecise but fast.

\paragraph{Relaxation Heuristic} \label{sec:heuristic} For our evaluation, we use the following heuristic, applied before each convolutional layer. The heuristic is parameterized by a relaxation percentage $p\in[0,1]$ and a clustering parameter $k\in\mathbb{N}$. Each chain of connected segments with $t>1000$ nodes is traversed in order, and each segment is turned into its bounding box, until the chain ends, the total number of different segment endpoints visited exceeds $t/k$ or we find a segment whose length is strictly above the $p$-th percentile, computed over all segment lengths in the chain prior to applying the heuristic. All bounding boxes generated in one such step (from adjacent segments) are then merged, the next segment (if any) is skipped, and the traversal is restarted on the remaining segments of the chain. This way, each chain is split into some chains and a number of new boxes.

\section{Certification of Probabilistic Properties}
We now extend the analysis to the probabilistic case and formally define \domain{} for probabilistic properties.

Our goal is to automatically show that images $\vx$ drawn from a given input distribution $\mu$ map to desirable outputs $\sD$ with a probability in some interval $[l,u]$. We can write this property as $\Pr_{\vx\sim\mu}[d(\vx)\in\sD]\in[l,u]$. For example, we can choose $d$ as a decoder, $\mu$ as the uniform distribution on a line segment in its latent space, $\sD$ as the set of images for which a given classifier detects a desired attribute and $[l,u]=[0.95,1]$. The property then states that for at least a fraction $0.95$ of the interpolated points, the given classifier detects the desired attribute.

Note that in contrast to the deterministic setting, probabilistic properties of this kind do not compose naturally. We therefore reformulate the property by defining sets $\sX$ and $\sY$ of \emph{probability distributions} and a distribution transformer $f$, in analogy to deterministic properties. Namely, we let $\sX=\{\mu\}$, \mbox{$\sY=\{\nu\mid \Pr_{\vy\sim\nu}[\vy\in\sD]\in[l,u]\}$} and $f=d_*$, where $d_*$ is the \emph{pushforward} of $d$. I.e., $f$ maps a distribution of inputs to $d$ to the corresponding distribution of outputs of $d$. With those definitions, our property again reads $f[\sX]\subseteq\sY$ and can be decomposed into properties $(L_i)_*[\sA_i]\subseteq \sA_{i+1}$ talking about each individual layer.

Therefore, we again overapproximate $\sX$ with $\sA_0$ such that $\sX\subseteq\sA_0$ and push it through each layer of the network, computing sets $\sA_1,\ldots,\sA_l$. Instead of activation vectors, the sets $\sA_i$ now contain \emph{distributions} over such vectors. It therefore remains to select a class of sets of distributions that we can represent using a few real parameters.

\paragraph{Lifting} Note that we can reuse any deterministic analysis directly as a probabilistic analysis, by ignoring probabilities. I.e., we let the set $\sA_i$ be the set of all probability distributions on a corresponding set $\sA_i'$ of (deterministic) activation vectors. The analysis then just propagates the deterministic sets $\sA_i'$ as it would in the deterministic setting. For example, if each set $\sA'_i$ is a box, $\sA_i$ is the set of all distributions on this box $\sA'_i$, and the analysis propagates the box constraints using interval arithmetic. Of course, such an analysis is rather limited, as it can at most prove properties with $l=0$ or $u=1$. For example, it would be impossible to prove that a probability is between $0.6$ and $0.8$ using only this kind of lifted analysis. However, interval arithmetic, lifted in this fashion, is a powerful component of \domain{} for probabilistic properties, detailed below.

\paragraph{Convex Combinations}
We can represent a set $\sA_i$ as a pointwise convex combination of sets of distributions $\sA_i^{(j)}$ with weights $\lambda_i^{(j)}$.
Formally, for fixed sets $\sA_i^{(j)}$ and weights $\lambda_i^{(j)}\ge0$ where $\sum_{j=1}^{k_i}\lambda_i^{(j)}=1$, the set $\sA_i$ contains all distributions of the form $\sum_{j=1}^{k_i} \lambda_i^{(j)}\cdot\mu^{(j)}$, where each $\mu^{(j)}$ is some distribution chosen from the corresponding set $\sA_i^{(j)}$.

For example, if $\sA_i^{(j)}$ is lifted from a box ${\sA'}_i^{(j)}$, then $\sA_i^{(j)}$ represents all distributions on the box ${\sA'}_i^{(j)}$. If those boxes are disjoint, the weight $\lambda_i^{(j)}$ is the probability of finding the neural network activations in layer $i$ within the box ${\sA'}_i^{(j)}$. In general, we can intuitively think of $\sA_i$ as making the assertion that there exists a certain random process producing results distributed like the activations in the $i$-th layer. The process first randomly selects one of the sets $\sA_i^{(j)}$ according to the probabilities $\lambda_i^{(j)}$ and then samples from a fixed distribution $\mu^{(j)}$ that is known to be in $\sA_i^{(j)}$. 

Similar to unions in the deterministic case, it is valid to propagate each set $\sA_i^{(j)}$ independently and to form the convex combination of the results using the same weights $\lambda_{i+1}^{(j)}=\lambda_i^{(j)}$ to obtain $\sA_{i+1}$.

\subsection{\domain{} for Probabilistic Properties}
Probabilistic \domain{} represents the sets $\sA_i$ as convex combinations of sets $\sA_i^{(j)}$, where each set $\sA_i^{(j)}$ is either a lifted box or contains a single distribution on a segment. 

Formally, this means $$\sA_i=\left\{\sum_{j=1}^{k_i}\lambda_i^{(j)}\cdot\mu^{(j)}\;\middle|\;\mu^{(1)}\in\sA_i^{(1)},\ldots,\mu^{(k_i)}\in\sA_i^{(k_i)}\right\},$$ where for each $j$, the set $\sA_i^{(j)}$ is either the set of distributions on a box $\prod_{l=1}^{n_i}[a_l,b_l]$ with given lower bounds $\va$ and upper bounds $\vb$, or $\sA_i^{(j)}=\{\nu\}$, where $\nu$ is a distribution on a segment $\overline{\vx_1\vx_2}$ with given endpoints $\vx_1$ and $\vx_2$ in $\R^{n_i}$. To automate the analysis, we can represent $\sA_i$ as a list of bounds of boxes with associated weights $\lambda_i^{(j)}$, as well as a list of segments with associated probability distributions and weights $\lambda_i^{(j)}$. If, as in our evaluation, we consider a restricted case, where probability distributions on segments are uniform, it suffices to associate a weight to each segment. The weights should be non-negative and sum up to $1$.

The set $\sA_i$ can be propagated through layer $L_i$ to obtain $\sA_{i+1}$ in a similar fashion to deterministic analysis. However, when splitting a segment, we now also need to split the distribution associated to it. For example, if we want to split the segment $\sL=\overline{\vc\vd}$ with distribution $\nu$ and weight $\lambda$ into two segments $\sL'=\overline{\vc\ve}$ and $\sL''=\overline{\ve\vd}$ with $\sL'\cup\sL''=\sL$, we have to form distributions $\nu', \nu''$ and weights $\lambda',\lambda''$ where $\lambda'=\lambda\cdot\Pr_{\vx\sim\nu}[\vx\in\sL']$, $\lambda''=\lambda\cdot\Pr_{\vx\sim\nu}[\vx\in\sL''\setminus\sL']$, $\nu'$ is $\nu$ conditioned on the event $\sL'$ and $\nu''$ is $\nu$ conditioned on the event $\sL''$. For example, if distributions on segments are uniform, this would result in the weight being split according to the relative lengths of the two new segments. A segment can be split into more than two segments by applying this procedure the required number of times. To propagate a lifted box, we apply interval arithmetic, preserving the box's weight. In practice, this is the same computation used for deterministic propagation of a box.

We focus on the case where we want to propagate a singleton set containing the uniform distribution on a segment $\sL=\overline{\vx_1\vx_2}$ through the neural network. In this case, each distribution on a propagated segment will still remain uniform, and it suffices to store a segment's weight without an explicit representation for the corresponding distribution, as noted above. As in the deterministic setting, if we apply the analysis to the uniform distribution on a segment without applying relaxation operators, the analysis will compute an exact representation of the output distribution. I.e., $\sA_l$ will be a singleton set containing the distribution of outputs obtained when the neural network is applied to inputs distributed uniformly at random on $\sL$.

\paragraph{Relaxation} As this does not scale, we again apply relaxation operators. Similar to the deterministic setting, we can replace a set of distributions $\sA_i$ by another set of distributions $\sA_i'$ with $\sA_i\subseteq \sA_i'$.

\paragraph{Relaxation Heuristic}
Here, we use the same heuristic described for the deterministic setting. When replacing a segment by its bounding box, we preserve its weight. When merging multiple boxes, their weights are added to give the weight for the resulting box.

\paragraph{Computing Bounds} Given the set $\sA_l$, describing the possible output distributions of the neural network, we want to compute optimal bounds on the robustness probabilities $\sP=\{\Pr_{\vy\sim\nu}[\vy\in\sD]\mid \nu\in\sA_l\}$. The part of the distribution tracked using segments has all its probability mass in perfectly determined locations, while the probability mass in each box can be located anywhere inside it.

We can therefore compute bounds as
\[
(l,u)=(\min\sP,\max\sP)=\left(e+\sum_{j\in\sL} \lambda_l^{(j)},e+\sum_{j\in\sU} \lambda_l^{(j)}\right),
\] 
where $e$ is the probability of finding the output on one of the segments. If $\sD$ is described as a set of linear constraints, we can compute $e$ by splitting the segments such that they do not cross the constraints and summing up all weights of resulting segments that are subsets of $\sD$. $\sL$ is the set of all indices of lifted boxes that are subsets of $\sD$, and $\sU$ is the set of all indices of lifted boxes that have a non-empty intersection with $\sD$.

\section{Evaluation}
\newcommand{\consistency}{\mathcal{C}}
\newcommand{\avgconsistency}{\hat{\mathcal{C}}}

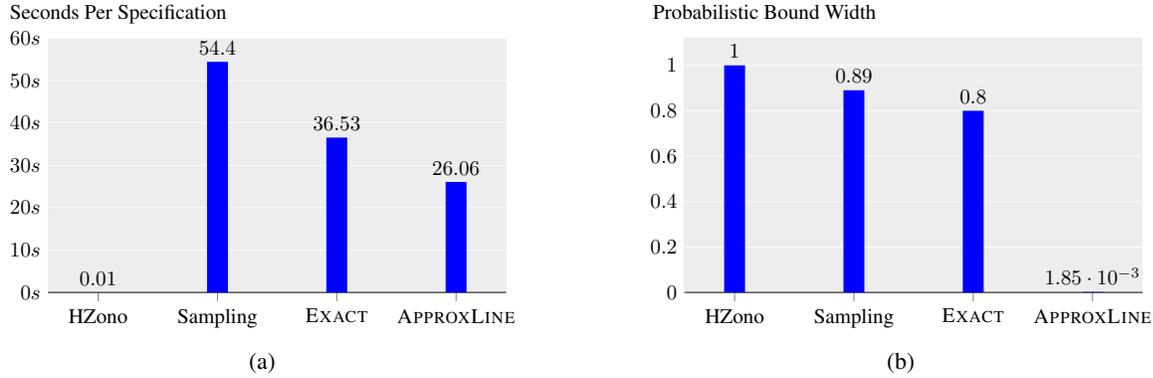
\begin{figure*}[t]
\hspace{-0.04\textwidth}%
\begin{subfigure}[b]{0.45\textwidth}
\centering
\begin{tikzpicture}[scale=0.79]\def\size{0.74}\begin{axis}[
  ybar,
  scale=\size,
  width=12cm,
  height=7.37cm,
  xlabel={},
  ylabel={Seconds Per Specification},
  symbolic x coords= {HZono, Sampling, ExactL, Approx},
  xticklabels={HZono, Sampling, \eline{}, \domain{}},
  ytick={0,10,...,60},
  nodes near coords,
  every node near coord/.append style={/pgf/number format/fixed},
  xtick=data,
  yticklabel=$\pgfmathprintnumber{\tick}s$,
  ymin=0, ymax=60,
  y axis line style={draw=none},
  axis background style = {fill=lightergrey},
  ylabel style={rotate=-90, at={(0.06,1.16)}, anchor=north west},
  y tick style={draw=none},
  legend cell align={left},
  legend style={draw=none, fill=none, at={(0.55,0.75)}, anchor=south west},
  xtick align=outside,
  axis y line*=left,
  axis x line*=bottom, 
  ymajorgrids=true,
  xmajorgrids=false,
  grid style={solid, draw=white},
  enlarge x limits=0.14,
]

\pgfplotstableread{plots/standard_dataset_metric/results_best.dat}\loadedtable;

\addplot[fill=blue,draw=none] table[
  x=name, 
  y=Speed] {\loadedtable};

\end{axis}\end{tikzpicture}
\caption{}
\end{subfigure}
\hspace{0.04\textwidth}%
\begin{subfigure}[b]{0.45\textwidth}
\centering
\begin{tikzpicture}[scale=0.79]\def\size{0.74}\begin{axis}[
  ybar,
  scale=\size,
  width=12cm,
  height=7.37cm,
  xlabel={},
  ylabel={Probabilistic Bound Width},
  symbolic x coords= {HZono, Sampling, ExactL, Approx},
  xticklabels={HZono, Sampling, \eline{}, \domain{}},
  nodes near coords,
  ytick={0,0.2,...,1},
  xtick=data,
  yticklabel=$\pgfmathprintnumber{\tick}$,
  ymin=0, ymax=1.12,
  y axis line style={draw=none},
  axis background style = {fill=lightergrey},
  ylabel style={rotate=-90, at={(0.075,1.16)}, anchor=north west},
  y tick style={draw=none},
  legend cell align={left},
  legend style={draw=none, fill=none, at={(0.55,0.75)}, anchor=south west},
  xtick align=outside,
  axis y line*=left,
  axis x line*=bottom, 
  ymajorgrids=true,
  xmajorgrids=false,
  grid style={solid, draw=white},
  enlarge x limits=0.14,
]

\pgfplotstableread{plots/standard_dataset_metric/results_best.dat}\loadedtable;

\addplot[fill=blue, draw=none] table[
  x=name, 
  y=VerifDiff_avg] {\loadedtable};

\end{axis}\end{tikzpicture}
\caption{}
\end{subfigure}
\vspace{-0.3cm}
\caption{A comparison of different methods that compute probabilistic bounds for $\avgconsistency$. \domain{} uses $p=.02$ and $k=100$.}
\label{best_graphs}
\end{figure*}

It is well known that certain generative models appear to produce interpretable transformations between outputs for interpolations of encodings in the latent space \citep{dumoulin2016adversarially, mathieu2016disentangling, bowman2015generating, radford2015unsupervised, mescheder2017adversarial, ha2017neural, dinh2016density, larsen2015autoencoding, van2016conditional, lu2018attribute, he2019attgan}. That is, as we move from one latent vector to another, there are interpretable attributes of the outputs that gradually appear or disappear. This leads to the following question: given encodings of two outputs with a number of \emph{shared} attributes, what fraction of the line segment between the encodings generates outputs sharing those same attributes? To answer this question, we can verify a generator using a trusted attribute detector, or we can verify an attribute detector based on a trusted generator. For both tasks, we have to analyze the outputs of neural networks restricted to segments.  We first describe robustness terminology.

Let $N \colon\R^m\to\R^n$ be a neural network with $m$ inputs and $n$ outputs which classifies $\vx\in\R^m$ to class $\argmax_i N(\vx)_i$.

\textbf{Specification} A robustness specification is a pair $( \sX,\sY )$ where $\sX\subseteq\R^m$ is a set of input activations and $\sY\subseteq\R^n$ is a set of permissible outputs for those inputs.

\textbf{Deterministic robustness} Given a specification $( \sX,\sY )$, a neural network $N$ is said to be $(\sX,\sY)$-robust if for all $\vx\in \sX$, we have $N(\vx)\in \sY$. In the adversarial robustness literature, $\sX$ is usually an $l_{2}$- or $l_{\infty}$-ball, and $\sY$ is a set of outputs  corresponding to a specific classification.  In our case, $\sX$ is a segment connecting two encodings.  The {\em deterministic verification problem} is to prove (ideally with $100\%$ confidence) that a network is robust for a specification. As deciding robustness is NP-hard \citep{katz2017reluplex}, to enable scalability, the problem is frequently relaxed so the verifier may sometimes say the network is not robust when it is (while still maintaining soundness: if the network is not robust, it will certainly flag it).

\textbf{Probabilistic robustness} Even if $N$ is not entirely robust, it may still be useful to quantify its lack of robustness. Given a distribution $\mu$ over $\sX$, we are interested in finding provable bounds on the {\em robustness probability} $\Pr_{\vx\sim\mu}[N(\vx)\in \sY]$, which we call {\em probabilistic [robustness] bounds}.

\subsection{Experimental Setup}

We refer to our verifier as \textsc{ApproxLine} (both deterministic and probabilistic versions) where the relaxation heuristic uses relaxation percentage $p$ and clustering parameter $k$. We implement \domain{} as in the DiffAI \cite{diffai} framework, taking advantage of the GPU parallelization provided by PyTorch \citep{paszke2017automatic}. Additionally, we use our implementation of \domain{} to compute exact results without approximation.
To obtain exact results, it suffices to set the relaxation percentage $p$ to $0$ in which case the clustering parameter $k$ can be ignored. We write \eline{} to mean verification using $\textsc{ApproxLine}^0_k$.  This should produce equivalent (up to floating point) results \textsc{ExactLine}, but is implemented on the GPU.
We run on a machine with a GeForce GTX 1080 with 12 GB of GPU memory, and four processors with a total of 64 GB of RAM.

\subsection{Generative specifications}

For generative specifications, we use autoencoders with either $32$ or $64$ latent dimensions trained in two different ways: \vae{} and \aedcyclegan{}, described below. We train them to reconstruct CelebA with image sizes $64\times64$.  We always use Adam \cite{kingma2014adam} with a learning rate of $0.0001$ and a batch size of $100$.  The specific network architectures are described in Appendix~\ref{archs}.  For CelebA, the decoder has $74128$ neurons and the attribute detector has $24676$.

\textbf{\vae{}$_l$} is a variational autoencoder \citep{vae} with $l$ latent dimensions.

\textbf{\aedcyclegan{}$_l$} is a repurposed CycleGAN \citep{cyclegan}.
While originally designed for style transfer between  distributions $P$ and $Q$, we use it as an autoencoder where the generator behaves like a GAN such that encodings are ideally distributed well in the latent space.  For the latent space distribution $P$ we use a normal distribution in $l$ dimensions with a feed forward network $D_P$ as its discriminator. The distribution $Q$ is the ground truth data distribution. For its discriminator $D_Q$ we use the BEGAN method \citep{began}, which determines an example's realism based on an autoencoder with $l$ latent dimensions, trained to reproduce $Q$ and adaptively to fail to reproduce the GAN generator's distribution.

\textbf{Attribute Detector}
is trained to recognize the $40$ attributes provided by CelebA.
The attribute detector has a linear output.
The attribute $i$ is detected in the image if the $i$-th component of the attribute detector's output is strictly positive.
It is trained using Adam for $300$ epochs.

\subsection{Speed and precision results}

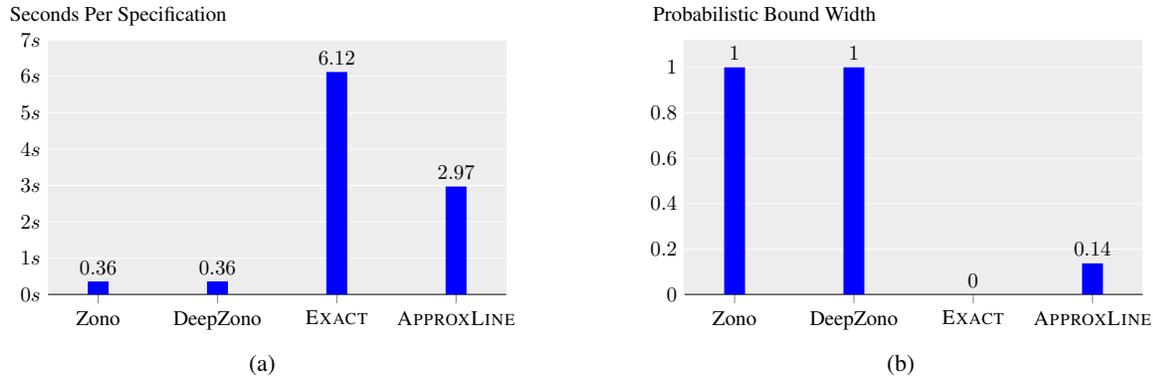
\begin{figure*}[t]
\hspace{-0.04\textwidth}%
\begin{subfigure}[b]{0.45\textwidth}
\centering
\begin{tikzpicture}[scale=0.79]\def\size{0.74}\begin{axis}[
  ybar,
  scale=\size,
  width=12cm,
  height=7.37cm,
  xlabel={},
  ylabel={Seconds Per Specification},
  symbolic x coords= {Zono, DeepZono, ExactL, Approx},
  xticklabels={Zono, DeepZono, \eline{}, \domain{}},
  ytick={0,1,...,7},
  nodes near coords,
  every node near coord/.append style={/pgf/number format/fixed},
  xtick=data,
  yticklabel=$\pgfmathprintnumber{\tick}s$,
  ymin=0, ymax=7,
  y axis line style={draw=none},
  axis background style = {fill=lightergrey},
  ylabel style={rotate=-90, at={(0.06,1.16)}, anchor=north west},
  y tick style={draw=none},
  legend cell align={left},
  legend style={draw=none, fill=none, at={(0.55,0.75)}, anchor=south west},
  xtick align=outside,
  axis y line*=left,
  axis x line*=bottom, 
  ymajorgrids=true,
  xmajorgrids=false,
  grid style={solid, draw=white},
  enlarge x limits=0.14,
]

\pgfplotstableread{plots/standard_dataset_metric/mnist_results_best.dat}\loadedtable;

\addplot[fill=blue,draw=none] table[
  x=name, 
  y=Speed] {\loadedtable};

\end{axis}\end{tikzpicture}
\caption{}
\end{subfigure}
\hspace{0.04\textwidth}%
\begin{subfigure}[b]{0.45\textwidth}
\centering
\begin{tikzpicture}[scale=0.79]\def\size{0.74}\begin{axis}[
  ybar,
  scale=\size,
  width=12cm,
  height=7.37cm,
  xlabel={},
  ylabel={Probabilistic Bound Width},
  symbolic x coords= {Zono, DeepZono, ExactL, Approx},
  xticklabels={Zono, DeepZono, \eline{}, \domain{}},
  nodes near coords,
  ytick={0,0.2,...,1},
  xtick=data,
  yticklabel=$\pgfmathprintnumber{\tick}$,
  ymin=0, ymax=1.12,
  y axis line style={draw=none},
  axis background style = {fill=lightergrey},
  ylabel style={rotate=-90, at={(0.075,1.16)}, anchor=north west},
  y tick style={draw=none},
  legend cell align={left},
  legend style={draw=none, fill=none, at={(0.55,0.75)}, anchor=south west},
  xtick align=outside,
  axis y line*=left,
  axis x line*=bottom, 
  ymajorgrids=true,
  xmajorgrids=false,
  grid style={solid, draw=white},
  enlarge x limits=0.14,
]

\pgfplotstableread{plots/standard_dataset_metric/mnist_results_best.dat}\loadedtable;

\addplot[fill=blue, draw=none] table[
  x=name, 
  y=VerifDiff_avg] {\loadedtable};

\end{axis}\end{tikzpicture}
\caption{}
\end{subfigure}
\vspace{-0.3cm}
\caption{A comparison of speed and accuracy of domains for computing $\avgconsistency$ on MNIST. \domain{} uses $p=.95$ and $k=25$.}
\label{mnist_speed_graphs}
\end{figure*}

Given a generative model capable of producing data manifold interpolations between inputs, there are many different verification goals one might pursue. For example, we can check whether the generative model is correct with respect to a trusted classifier or whether a classifier is robust to interpretable interpolations produced by a trusted generator. Even trusting neither generator nor classifier, we might wish to verify their mutual consistency.

We address all of these goals by efficiently computing the \textit{attribute consistency} of a generative model with respect to an attribute detector:
for a point picked uniformly between the encodings $\ve_1$ and $\ve_2$ of ground truth
inputs both with or without matching attributes $i$, we would like to determine the probability that its decoding will have (or not) the same attribute.
We define attribute consistency as
\begin{equation*}
\consistency_{i,n_A, n_D}(\ve_1, \ve_2) = \Pr_{\ve \sim U(\overline{\ve_1\ve_2})}[\sign n_A(n_D(\ve))_i = t],
\end{equation*}
where $t = \sign n_A(n_D(\ve_1))_i$.
We will frequently omit the attribute detector $n_A$ and the decoder $n_D$ from $\consistency$ if it is clear from context which networks are being evaluated.

Here, we demonstrate that probabilistic \domain{} is precise and efficient enough to provide useful bounds on attribute consistency for interesting generators and specifications. Specifically, we compare \domain{} to a variety of other methods for providing probabilistic bounds.
We do this first for \aedcyclegan{}$_{32}$ trained for $200$ epochs.

Suppose $P$ is a set of unordered pairs $\{\va,\vb\}$ from the data with $\sign \va_{A,i} = \sign \vb_{A,i}$ for every attribute, where $\va_{A,i}$ is label of attribute $i$ for $\va$.
Using each method, we find bounds on the true value of \textit{average attribute consistency} as
$\avgconsistency_P = \text{mean}_{\va,\vb \in P, i} \consistency_{i, n_A, n_D}( n_E(\va), n_E(\vb))$ where $n_E$ is the encoding network.
Each method finds a probabilistic bound, $[l,u]$, such that $l \leq \avgconsistency \leq u$.
We call $u-l$ its width.

\begin{figure*}[t]
\floatsetup[subfigure]{capbesideposition={left,center}}
\ffigbox{%
\begin{subfloatrow}[1]
      \fcapside[\FBwidth]
        {\caption{}\label{real_spec}}%
        {\includegraphics[width=0.828\textwidth]{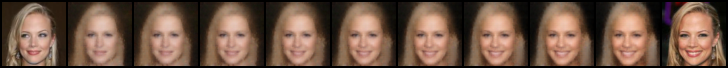}%
          \hspace{45pt}}%
\end{subfloatrow}\vskip2pt
\begin{subfloatrow}[1]
      \fcapside[\FBwidth]
        {\caption{}\label{flip_spec}}%
        {\includegraphics[width=0.828\textwidth]{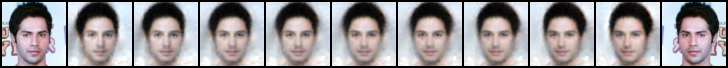}%
         \hspace{45pt}}%
\end{subfloatrow}
\vskip2pt\begin{subfloatrow}[1]
      \fcapside[\FBwidth]
        {\caption{}\label{attribute_spec}}%
        {\includegraphics[width=0.752724\textwidth]{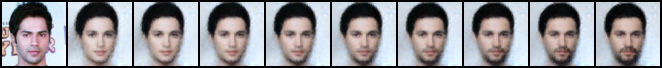}%
          \hspace{0.07527267\textwidth}\hspace{45pt}}%
\end{subfloatrow}
}{
\vspace{-0.5cm}
\caption{Examples of interpolated images, with originals at the 1st and 11th positions.
Their immediate neighbors are their reconstructions.
(a) is between the same person with the same attributes using \aedcyclegan{}$_{32}$ after $200$ epochs.
(b) is between horizontally flipped images using \aedcyclegan{}$_{64}$ after $20$ epochs.
(c) is between an image and the addition of the ``mustache'' feature vector using \aedcyclegan{}$_{32}$ again.
}
}
\end{figure*}
\begin{figure*}[t]
\floatbox[{\capbeside\thisfloatsetup{capbesideposition={left,top},capbesidewidth=0.45\linewidth}}]{figure}[\FBwidth]
{\caption{
Comparing the different generative models using probabilistic \domain{} with $p=0.02$ and $k=200$ to provide lower bounds for $\text{mean}_{i}\consistency_i(n_E(\va),n_E(\text{Flipped}(\va)))$, where $\va$ and $\text{Flipped}(\va)$ are the images shown in \fig{flip_spec}.
The width of the largest probabilistic bound was smaller than $3\times 10^{-6}$, so only the lower bounds are shown. Fewer than $50$ seconds were necessary to compute each bound, and the fastest computation was for \aedcyclegan{}$_{64}$ at $30$ seconds.}\label{compare_flip}}
{
\begin{tikzpicture}[scale=0.8]\def\size{0.75}\begin{axis}[
  ybar,
  scale=\size,
  width=12.5cm,
  height=6.5cm,
  xlabel={},
  ylabel={Lower Bound on Correctness},
  symbolic x coords= {VAE$_{32}$, AEDCycleGAN$_{32}$, VAE$_{64}$, AEDCycleGAN$_{64}$},
  xticklabels={\vae{}$_{32}$, \aedcyclegan{}$_{32}$, \vae{}$_{64}$, \aedcyclegan$_{64}$},
  xtick=data,
  yticklabel=$\pgfmathprintnumber{\tick}$,
  ytick={0.75,0.8,0.85,0.9, 0.95, 1},
  nodes near coords,
  every node near coord/.append style={/pgf/number format/fixed},
  ymin=0.75, ymax=1,
  y axis line style={draw=none},
  axis background style = {fill=lightergrey},
  ylabel style={rotate=-90, at={(0.08,1.23)}, anchor=north west},
  y tick style={draw=none},
  legend cell align={left},
  legend style={draw=none, fill=none, at={(0.55,0.75)}, anchor=south west},
  xtick align=outside,
  axis y line*=left,
  axis x line*=bottom, 
  ymajorgrids=true,
  xmajorgrids=false,
  grid style={solid, draw=white},
  enlarge x limits=0.08,
]

\pgfplotstableread{plots/compare_models/flip/results.dat}\loadedtable;

\addplot[fill=blue, draw=none] table[
  x=name, 
  y=LB_avg] {\loadedtable};

\end{axis}\end{tikzpicture}
}
\end{figure*}

We compare probabilistic \domain{} against two other probabilistic analyses: \eline{} (\textsc{ApproxLine} with $p=0$), and HZono \citep{diffai} lifted probabilistically. We also compare against probabilistic sampling with binomial confidence intervals on $\consistency$.

For probabilistic sampling, we take samples and recalculate the Clopper-Pearson interval with a confidence of $99.99\%$ until the interval width is below $.002$ (chosen to be the same as our best result with \domain{}).
To avoid incorrect calculation, we discard this result, and resample using the estimated number of samples.
Notably, the probabilistic bound returned by the analyses is guaranteed to be correct $100\%$ of the time, while for sampling it is only guaranteed to be correct $99.99\%$ of the time.

For all methods, we report a bound of $[0,1]$ in the case of either out-of-memory error, or 60s timeout.
For \domain{}, if an out-of-memory error occurs, we refine the hyperparameters using schedule A in Appendix~\ref{schedule} and restart (without resetting the timeout clock).
\fig{best_graphs} shows the results of running these on $\lvert P\rvert=100$ pairs of matching celebrities with matching attribute labels, chosen uniformly at random from CelebA (each method uses the same $P$).

While HZono is the fastest method, it is unable to prove any specifications.
Sampling and \eline{} appear to be similar in speed to \domain{}, but
the widths of the probabilistic bounds they produce is much larger.
This is because Sampling frequently times out, and \eline{} frequently exhausts GPU memory.
On the other hand, \domain{} provides an average probabilistic bound width of less than $0.002$ in under $30$s with perfect confidence (compared with the lower confidence provided by sampling).

\paragraph{Comparing to Precise Convex Abstract Domains}

Here we compare to two precise convex relaxation methods: Zonotope \cite{ai2} and DeepZono \cite{eran}.  At the expense of memory and speed, these domains are able to verify more properties.  Due to increased memory requirements, we were unable to use them on the GPU for the CelebA networks described.
In order to demonstrate their inadequacy for line specifications in generative networks, we adapted the same networks for MNIST \cite{mnist} such that it could still be analyzed on the GPU with the DiffAI implementation of Zonotope and DeepZono (from \citet{eran}).  We trained a \vae{}$_{50}$ for 10 epochs with a learning rate of $0.001$ with the networks described in Appendix~\ref{mnist_archs}.  The decoder has $74128$ neurons and the attribute detector has $23820$ neurons.  Here, the attribute detector works similarly, but attribute $i$ is considered present for images of the digit $i$. 

We analyzed 100 pairs with Zonotope, DeepZono, and ApproxLine with $p=0.95$ and $k=25$, and \eline{}. No timeouts occurred.  The results are shown in \fig{mnist_speed_graphs}.  Despite their speed, the precision of Zonotope and DeepZono is not enough to prove anything. While \eline{} is able to perfectly prove since it no longer runs out of GPU memory, it takes more than twice as long as \domain{}.

\subsection{Use cases for \domain{}}

We now demonstrate how to check the attribute consistency of a model against an attribute detector.
We do this for two possible generative specifications: (i) generating rotated heads using flipped images,
and (ii) adding previously absent attributes to faces.
For the results in this section, we use schedule B described in Appendix~\ref{schedule}.

\paragraph{Comparing models with turning heads}
It is known that VAEs are capable of generating images with intermediate poses from flipped images.
An example of this transformation is shown in \fig{flip_spec}.
Here, we show an example of using \domain{} to compare the effectiveness of different autoencoding models for this task.
We trained the $4$ models described above for $20$ epochs,
and created a specification between encodings of the flipped images shown in \fig{flip_spec}.
For a head that is turned in one direction,
ideally the different reconstructions will correspond to images of different orientations of the head in 3D space.
As none of the CelebA attributes correspond to pose, the attribute detector should recognize the same set of attributes for all interpolations.
We used deterministic \domain{} with $p=0.02$ and $k=200$ to demonstrate which attributes provably remain 
correct for every possible interpolation (as visualized in Appendix~\ref{sec:deterministic_flip}).  While we are able to show in the worst case, 32 out of 40 attributes are entirely robust to flipping, some attributes are not robust across interpolation.
\fig{compare_flip}
demonstrates the results of using probabilistic \domain{} to find the average lower bound on the fraction of the input interpolation encodings which do result in the correct attribute appearing in the output image.

\paragraph{Verifying attribute independence}
We now use \domain() to demonstrate that attribute detection for one feature is invariant to a transformation of an independent feature.
Specifically, we verify for a single image the effect of adding a mustache. This transformation is shown in \fig{attribute_spec}.
To do this, we find the attribute vector $m$ for ``mustache'' ($i=22$ in CelebA) using the 80k training-set images in the manner described by \cite{larsen2015autoencoding},
and compute probabilistic bounds for $\consistency_{j}(n_E(\vo), n_E(\vo) + 2m, o_{A, j})$ for $j\neq 22$ and the image $\vo$.
Using \domain{} we are able to prove that $30$ out of the $40$ attributes are entirely robust through the addition of a mustache.
Among the attributes which can be proven to be robust are $i=4$ for ``bald'' and $i=39$ for ``young''.
We are able to find that attribute $i=24$ for ``NoBeard'' is not entirely robust to the addition of the mustache vector.
We find a lower bound on the robustness probability of $0.83522$ and an upper bound of $0.83528$.

\section{Conclusion}

We presented a scalable non-convex relaxation to verify neural network properties restricted to line segments. Our method supports both deterministic and probabilistic certification and is able to verify, for the first time, interesting visual transformation properties based on latent space interpolation, beyond the reach of prior work.

\bibliography{ms}
\bibliographystyle{icml2020}

\clearpage
\appendix

\section{\domain{} Propogation Pseudocode}
\label{alg}

\begin{algorithm}[!h]
   \caption{Pseudocode for inference with \domain{}}
   \label{alg:approxline}
\begin{algorithmic}
   \STATE {\bfseries Input:} $k$ network layers with weights and biases $M_i, B_i$, and a line segment $a,b$ in the input space.
   \STATE {\bfseries Output:} $D$ a list of boxes and line segments describing possible regions of the output space with associated probabilities.

   \STATE $D = [ ($Segment$, 1, a,b) ] $.
   \FOR{$l=1$ {\bfseries to} $k-1$}
       \STATE $\tilde{D} = []$
       \FOR{$i=1$ {\bfseries to} $|D|$}
           \IF{$D_i == $ Segment }
              \STATE $a = D_{i,3}M_l + B_l$
              \STATE $b = D_{i,4}M_l + B_l$
              \STATE $T = [0,1]$
              \FOR{$d=1$ {\bfseries to} $|b_l|$}
                  \IF{$a_d < 0 \wedge B_d \geq 0$}
                     \STATE $T.$push$(\frac{ -a_d }{ | (a - b)_d | })$
                  \ELSIF{$a_d \geq 0 \wedge B_d < 0$}
                     \STATE $T.$push$(1 - \frac{ -b_d }{ | (a - b)_d | })$
                  \ENDIF
               \ENDFOR
               \STATE $T$.sort$()$
              \FOR{$t=1$ {\bfseries to} $|T|$}
                  \STATE $p = T_{t} - T_{t - 1}$
                  \STATE $\tilde{a} = (b - a) * T_{t - 1} + a$
                  \STATE $\tilde{b} = (b - a) * T_{t} + a$
                  \STATE $\tilde{D}.$push$(($Segment$, D_{i,2} * p, \tilde{a}, \tilde{b}))$
              \ENDFOR
           \ELSE
              \STATE $c = D_{i,3}M_l + b_l$ 
              \STATE $r = D_{i,4} |M_l|_p$
              \STATE $\tilde{c} = ReLU(c + r) + ReLU(c - r)$
              \STATE $\tilde{r} = ReLU(c + r) - ReLU(c - r)$
              \STATE $\tilde{D}.$push$(($Box$, D_{i,2}, 0.5 * \tilde{c}, 0.5 * \tilde{r}))$ 
           \ENDIF
       \ENDFOR
       \STATE $\tilde{P} = $Relax$(\tilde{D})$
       \FOR{$i=1$ {\bfseries to} $|\tilde{D}|$}
           \FOR{$p=1$ {\bfseries to} $|\tilde{P}|$}
              \IF{$\gamma(\tilde{D}_i) \subseteq \gamma(\tilde{P}_p)$ }  %
                  \STATE $\tilde{P}_{p,2} = \tilde{P}_{p,2} + \tilde{D}_{i,2}$
                  \STATE {\bf{delete}} $D_i$
                  \STATE {\bf{break}}
              \ENDIF
           \ENDFOR
       \ENDFOR
       \STATE $D = \tilde{D} + \tilde{P}$
   \ENDFOR
   \STATE {\bf{return}} $D$
\end{algorithmic}
\end{algorithm}

\clearpage
\section{Network Architectures}  
\label{archs}

For both models, we use the same encoders and decoders (even in the autoencoder descriminator from BEGAN), and always use the same attribute detectors.  
Here we use $\text{Conv}_{s} C \times W \times H$ to denote a convolution which produces
$C$ channels, with a kernel width of $W$ pixels and height of $H$, with a stride of $s$ and padding of $1$. $\text{FC}\;n$ is a fully connected layer which outputs $n$ neurons.  
$\text{ConvT}_{s,p} C\times W\times H$ is a transposed convolutional layer \citep{transposedconv} with a kernel width and height of $W$ and $H$ respectively and a stride of $s$ and padding of $1$ and out-padding of $p$, which produces $C$ output channels.

\begin{itemize}
\item \emph{Latent Descriminator} is a fully connected feed forward network with 5 hidden layers each of 100 dimensions.
\item \emph{Encoder} is a standard convolutional neural network:
\[
\begin{array}{r@{\hspace{5pt}}lr}
x & \to \text{Conv}_{1} 32\times3\times3 & \to \text{ReLU} \\
  & \to \text{Conv}_{2} 32\times4\times4 & \to \text{ReLU} \\
  & \to \text{Conv}_{1} 64\times3\times3 & \to \text{ReLU} \\
  & \to \text{Conv}_{2} 64\times4\times4 & \to \text{ReLU} \\
  & \to \text{FC}\;512 & \to \text{ReLU} \\
  & \to \text{FC}\;512 & \\
  & \to l. &
\end{array}
\]
\item \emph{Decoder} is a transposed convolutional network which has 74128 neurons:
\[
\begin{array}{r@{\hspace{5pt}}lr}
l & \to \text{FC}\;400 & \to \text{ReLU} \\
  & \to \text{FC}\;2048 & \to \text{ReLU} \\
  & \to \text{ConvT}_{2,1} 16\times3\times3 & \to \text{ReLU} \\
  & \to \text{ConvT}_{1,0} 3\times3\times3 & \\
  & \to x. &
\end{array}
\]
\item \emph{Attribute Detector} is a convolutional network which has 24676 neurons:
\[
\begin{array}{r@{\hspace{5pt}}lr}
x & \to \text{Conv}_{2} 16\times4\times4 & \to \text{ReLU} \\
  & \to \text{Conv}_{2} 32\times4\times4 & \to \text{ReLU} \\
  & \to \text{FC}\;100 & \\
  & \to 40. & 
\end{array}
\]
\end{itemize}

\section{MNIST Network Architectures}  
\label{mnist_archs}

Here we list the architectures for MNIST models.

\begin{itemize}
\item \emph{Encoder} is a standard convolutional neural network:
\[
\begin{array}{r@{\hspace{5pt}}lr}
x & \to \text{Conv}_{1} 32\times3\times3 & \to \text{ReLU} \\
  & \to \text{Conv}_{2} 32\times4\times4 & \to \text{ReLU} \\
  & \to \text{Conv}_{1} 64\times3\times3 & \to \text{ReLU} \\
  & \to \text{Conv}_{2} 64\times4\times4 & \to \text{ReLU} \\
  & \to \text{FC}\;512 & \to \text{ReLU} \\
  & \to \text{FC}\;512 & \\
  & \to l. &
\end{array}
\]
\item \emph{Decoder} is a transposed convolutional network which has 14512 neurons:
\[
\begin{array}{r@{\hspace{5pt}}lr}
l & \to \text{FC}\;400 & \to \text{ReLU} \\
  & \to \text{FC}\;1568 & \to \text{ReLU} \\
  & \to \text{ConvT}_{2,1} 16\times3\times3 & \to \text{ReLU} \\
  & \to \text{ConvT}_{1,0} 3\times3\times3 & \\
  & \to x. &
\end{array}
\]
\item \emph{Attribute Detector} is a convolutional network which has 4804 neurons:
\[
\begin{array}{r@{\hspace{5pt}}lr}
x & \to \text{Conv}_{2} 16\times4\times4 & \to \text{ReLU} \\
  & \to \text{Conv}_{2} 32\times4\times4 & \to \text{ReLU} \\
  & \to \text{FC}\;100 & \\
  & \to 10. & 
\end{array}
\]
\end{itemize}

\section{\domain{} Refinement Schedule}
\label{schedule}

While many refinement schemes start with an imprecise approximation and progressively tighten it, we observe that being only occasionally memory limited and rarely time limited, it conserves more time to start with the most precise approximation we have determined usually works, and progressively try less precise approximations as we determine that more precise ones can not fit into GPU memory. Thus, we start searching for a probabilistic robustness bound with \domain{}$^p_{N}$ and if we run out of memory, try \domain{}$^{\text{min}(1.5p,1)}_{\text{max}(0.95N, 5)}$ for schedule A, and \domain{}$^{\text{min}(3p,1)}_{\text{max}(0.95N, 5)}$ for schedule B.  This procedure is repeated until a solution is found, or time has run out.

\section{Effect of Approximation Parameters on Speed and Precision}

\begin{figure*}[h]
\centering
\begin{subfigure}[b]{0.48\textwidth}
\centering
\begin{tikzpicture}[scale=0.8]\def\size{0.75}\begin{axis}[
  ybar,
  scale=\size,
  width=14cm,
  height=12cm,
  xlabel={},
  ylabel={Time},
  symbolic x coords={$_{100}^{.02}$, $_{75}^{.02}$, $_{50}^{.02}$, $_{100}^{.3467}$, $_{75}^{.3467}$, $_{50}^{.3467}$, $_{100}^{.6733}$, $_{75}^{.6733}$, $_{50}^{.6733}$, $_{100}^{1}$, $_{75}^{1}$, $_{50}^{1}$},
  xtick=data,
  ytick={0,3,...,30},
  yticklabel=$\pgfmathprintnumber{\tick}s$,
  ymin=0, ymax=30,
  y axis line style={draw=none},
  axis background style = {fill=lightergrey},
  ylabel style={rotate=-90, at={(0.05,1.08)}, anchor=north west},
  y tick style={draw=none},
  legend cell align={left},
  legend style={draw=none, fill=none, at={(0.55,0.75)}, anchor=south west},
  xtick align=outside,
  axis y line*=left,
  axis x line*=bottom, 
  ymajorgrids=true,
  xmajorgrids=false,
  grid style={solid, draw=white},
  enlarge x limits=0.03,
]

\pgfplotstableread{plots/standard_dataset_metric/results_compare.dat}\loadedtable;

\addplot[fill=blue,draw=none] table[
  x=name, 
  y=Speed] {\loadedtable};

\end{axis}\end{tikzpicture}
\caption{}
\end{subfigure}
\hspace{0.03\textwidth}%
\begin{subfigure}[b]{0.48\textwidth}
\centering
\begin{tikzpicture}[scale=0.8]\def\size{0.75}\begin{axis}[
  ybar,
  scale=\size,
  width=14cm,
  height=12cm,
  xlabel={},
  ylabel={Prob Bound Width},
  symbolic x coords= {$_{100}^{.02}$, $_{75}^{.02}$, $_{50}^{.02}$, $_{100}^{.3467}$, $_{75}^{.3467}$, $_{50}^{.3467}$, $_{100}^{.6733}$, $_{75}^{.6733}$, $_{50}^{.6733}$, $_{100}^{1}$, $_{75}^{1}$, $_{50}^{1}$},
  xtick=data,
  yticklabel=$\pgfmathprintnumber{\tick}$,
  ytick={0,0.1,...,1.1},
  ymin=0, ymax=1,
  y axis line style={draw=none},
  axis background style = {fill=lightergrey},
  ylabel style={rotate=-90, at={(0.05,1.08)}, anchor=north west},
  y tick style={draw=none},
  legend cell align={left},
  legend style={draw=none, fill=none, at={(0.55,0.75)}, anchor=south west},
  xtick align=outside,
  axis y line*=left,
  axis x line*=bottom, 
  ymajorgrids=true,
  xmajorgrids=false,
  grid style={solid, draw=white},
  enlarge x limits=0.03,
]

\pgfplotstableread{plots/standard_dataset_metric/results_compare.dat}\loadedtable;

\addplot[fill=blue, draw=none] table[
  x=name, 
  y=VerifDiff_avg] {\loadedtable};

\end{axis}\end{tikzpicture}
\caption{}
\end{subfigure}
\caption{A comparison of speed (a) and Probabilistic Bound Widths (b) of \domain{} for different approximation hyperparameters, on 
\aedcyclegan{}$_64$ trained for 200 epochs.
}
\label{self_compare_graphs}
\end{figure*}

Here we demonstrate how modifying the approximation parameters, $p$ and $N$ of \domain{}$^{p}_{N}$ effect its speed and precision.  \fig{self_compare_graphs} shows the result of varying these on x-axis.  The bottom number, $N$ is the number of clusters that will be ideally made, and the top number $p$ is the percentage of nodes which are permitted to be clustered.

\section{Comparing the Deterministic Binary Robustness of Different Models}
\label{sec:deterministic_flip}

\begin{figure*}[h]
\centering
\newcommand{\scc}[1]{&\cellcolor{blue!30}}
\newcommand{\fal}[1]{&\cellcolor{red!50}}
\begin{tabular}{p{0.04cm}@{\hspace{-0.041cm}}p{0.04cm}@{};{0.4mm/0.6mm}p{0.04cm}@{};{0.4mm/0.6mm}p{0.04cm}@{};{0.4mm/0.6mm}p{0.04cm}@{};{0.4mm/0.6mm}p{0.04cm}@{};{0.4mm/0.6mm}p{0.04cm}@{};{0.4mm/0.6mm}p{0.04cm}@{};{0.4mm/0.6mm}p{0.04cm}@{};{0.4mm/0.6mm}p{0.04cm}@{};{0.4mm/0.6mm}p{0.04cm}@{};{0.4mm/0.6mm}p{0.04cm}@{};{0.4mm/0.6mm}p{0.04cm}@{};{0.4mm/0.6mm}p{0.04cm}@{};{0.4mm/0.6mm}p{0.04cm}@{};{0.4mm/0.6mm}p{0.04cm}@{};{0.4mm/0.6mm}p{0.04cm}@{};{0.4mm/0.6mm}p{0.04cm}@{};{0.4mm/0.6mm}p{0.04cm}@{};{0.4mm/0.6mm}p{0.04cm}@{};{0.4mm/0.6mm}p{0.04cm}@{};{0.4mm/0.6mm}p{0.04cm}@{};{0.4mm/0.6mm}p{0.04cm}@{};{0.4mm/0.6mm}p{0.04cm}@{};{0.4mm/0.6mm}p{0.04cm}@{};{0.4mm/0.6mm}p{0.04cm}@{};{0.4mm/0.6mm}p{0.04cm}@{};{0.4mm/0.6mm}p{0.04cm}@{};{0.4mm/0.6mm}p{0.04cm}@{};{0.4mm/0.6mm}p{0.04cm}@{};{0.4mm/0.6mm}p{0.04cm}@{};{0.4mm/0.6mm}p{0.04cm}@{};{0.4mm/0.6mm}p{0.04cm}@{};{0.4mm/0.6mm}p{0.04cm}@{};{0.4mm/0.6mm}p{0.04cm}@{};{0.4mm/0.6mm}p{0.04cm}@{};{0.4mm/0.6mm}p{0.04cm}@{};{0.4mm/0.6mm}p{0.04cm}@{};{0.4mm/0.6mm}p{0.04cm}@{};{0.4mm/0.6mm}p{0.04cm}@{};{0.4mm/0.6mm}p{0.04cm}@{}!{\color{white!0}\vrule}l}
\fal{}\scc{}\scc{}\scc{}\scc{}\scc{}\fal{}\scc{}\scc{}\scc{}\scc{}\scc{}\fal{}\scc{}\scc{}\scc{}\scc{}\scc{}\scc{}\scc{}\scc{}\scc{}\scc{}\scc{}\fal{}\fal{}\fal{}\fal{}\scc{}\scc{}\scc{}\scc{}\fal{}\scc{}\scc{}\scc{}\scc{}\scc{}\scc{}\scc{}&\cellcolor{white}\vae{}$_{32}$\\ \hline
\fal{}\scc{}\scc{}\scc{}\scc{}\scc{}\fal{}\scc{}\scc{}\scc{}\scc{}\scc{}\fal{}\scc{}\scc{}\scc{}\scc{}\scc{}\scc{}\scc{}\fal{}\scc{}\scc{}\scc{}\fal{}\scc{}\fal{}\fal{}\scc{}\scc{}\scc{}\scc{}\fal{}\scc{}\scc{}\scc{}\scc{}\scc{}\scc{}\scc{}&\cellcolor{white}\aedcyclegan{}$_{32}$ \\ \hline
\fal{}\scc{}\scc{}\scc{}\scc{}\scc{}\fal{}\scc{}\scc{}\scc{}\scc{}\scc{}\fal{}\scc{}\scc{}\scc{}\scc{}\scc{}\scc{}\scc{}\scc{}\scc{}\scc{}\scc{}\fal{}\fal{}\fal{}\fal{}\scc{}\scc{}\scc{}\scc{}\fal{}\scc{}\scc{}\scc{}\scc{}\scc{}\scc{}\scc{}&\cellcolor{white}\vae{}$_{64}$ \\ \hline
\scc{}\scc{}\scc{}\scc{}\scc{}\scc{}\fal{}\scc{}\scc{}\scc{}\scc{}\scc{}\fal{}\scc{}\scc{}\scc{}\scc{}\scc{}\scc{}\scc{}\scc{}\scc{}\scc{}\scc{}\scc{}\scc{}\fal{}\fal{}\scc{}\scc{}\scc{}\scc{}\scc{}\scc{}\scc{}\scc{}\scc{}\scc{}\scc{}\scc{}&\cellcolor{white}\aedcyclegan{}$_{64}$
\end{tabular}
\caption{
Blue means that the interpolative specification visualized in \fig{flip_spec} has been deterministically and entirely verified for the attribute (horizontal) using \domain{}$^{0.02}_{200}$.
Red means that the attribute can not be verified.
In all cases, this is because the specification was not robust for the attribute.
One can observe that the most successful autoencoder is \aedcyclegan{}$_{64}$.}
\label{deterministic_flip}
\end{figure*}

\fig{deterministic_flip} uses deterministic \domain{}$^{0.02}_{200}$ to demonstrate which attributes provably remain the
same and are correct (in blue) for every possible interpolation.

\end{document}